\ifdefined\pdfminorversion\pdfminorversion=4\fi
\documentclass[letterpaper,10pt,conference]{ieeeconf}
\IEEEoverridecommandlockouts
\usepackage{graphicx}
\usepackage{amsmath,amssymb}
\usepackage{booktabs,multirow,array}
\usepackage{xcolor}
\usepackage{cite}
\usepackage[hidelinks]{hyperref}
\usepackage{tikz}
\usetikzlibrary{arrows.meta,positioning,calc,fit,backgrounds}
\graphicspath{{./figures/figures_all/}}
\definecolor{traceblue}{RGB}{36,94,140}
\definecolor{traceteal}{RGB}{34,139,135}
\newcommand{\method}{\textsc{ACG-WAM}}
\newcommand{\objective}{\textsc{ACG-JEPA}}

\newcommand{\sg}{\operatorname{sg}}
\newcommand{\LN}{\operatorname{LN}}

\title{\method: World-Action Modeling\\
via Action-Conditioned Geometric Latent Prediction}

\author{
    \textbf{Jiangtao Liu}$^{1,2*}$\quad
    \textbf{Zishang Xiang}$^{1*}$\quad
    \textbf{Yage He}$^{2}$\quad
    \textbf{Lingguo Cui}$^{1}$\\
    \textbf{Baihai Zhang}$^{1}$\quad
    \textbf{Runqi Chai}$^{1}$\quad
    \textbf{Senchun Chai}$^{1\dag}$ \vspace{0.1cm}\\
    $^1$School of Automation, Beijing Institute of Technology\quad
    $^2$LimX Dynamics\vspace{0.05cm}\\
    \small $^*$Equal contribution.
    $^\dag$Corresponding author:chaisc97@bit.edu.cn
}

\IEEEaftertitletext{%
\vspace{-0.4ex}%
\begin{center}%
\includegraphics[width=\textwidth,trim=39bp 212bp 290bp 64bp,clip]{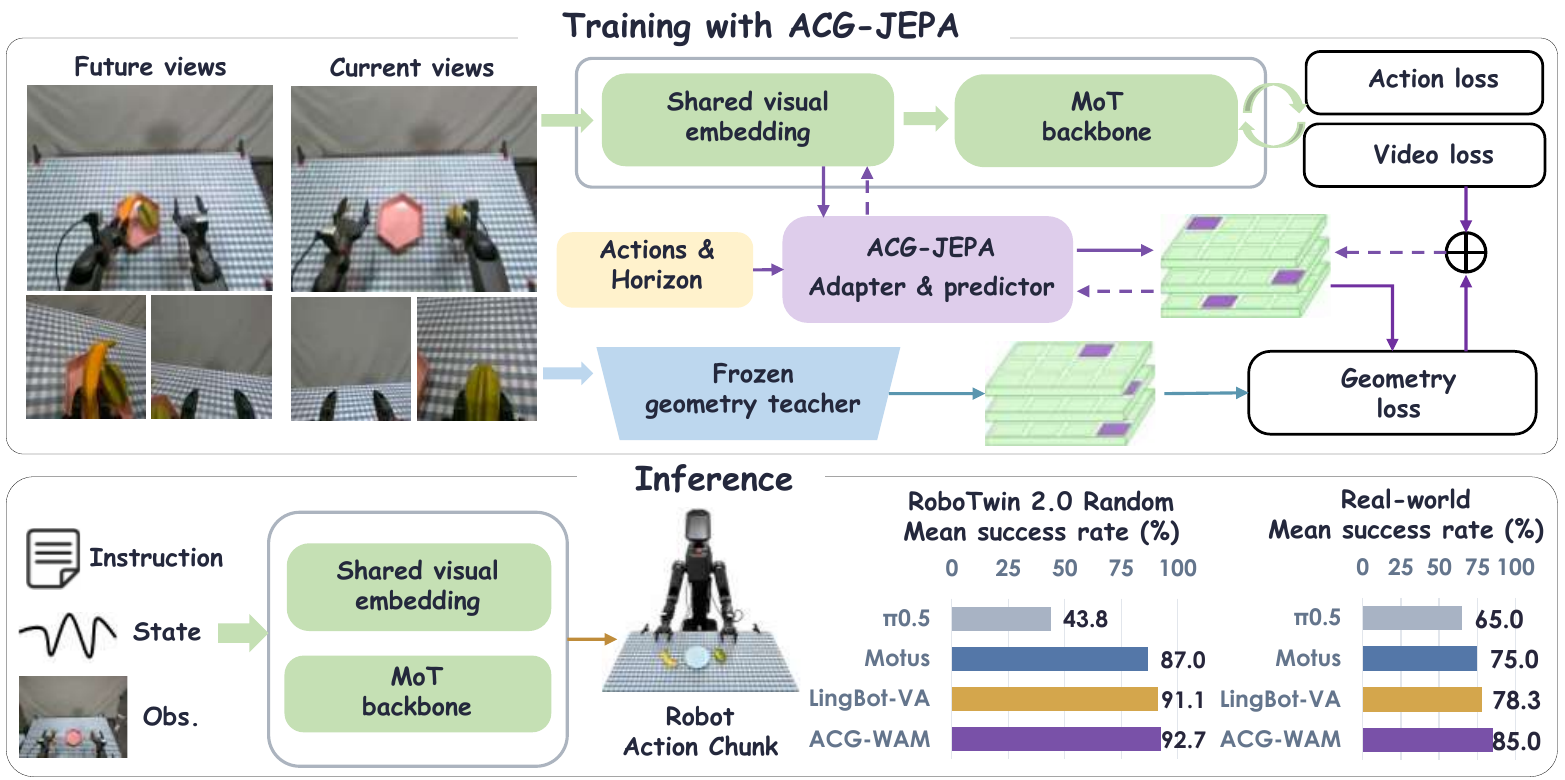}%
\end{center}%
\vspace{0.4ex}%
\refstepcounter{figure}\label{fig:teaser}%
\noindent\footnotesize
Fig.~\thefigure. Overview of \method{}. Visual Geometry Grounded Transformer (VGGT) supplies geometric targets
for prediction from current visual features and actions. The auxiliary loss updates
the shared visual embedding during training; inference uses the trained backbone.
\\[-0.3ex]
}
\begin{document}
\bstctlcite{ACG:BSTcontrol}
\maketitle
\thispagestyle{empty}
\pagestyle{empty}

\begin{abstract}
World action models jointly learn visual prediction and robot actions,
providing a way to use observations of scene evolution for policy learning.
Their video and action losses, however, provide no explicit target for
the geometric consequences of a demonstrated action sequence.
Moreover, visual features taken after temporal attention can contain future
observations, making them unsuitable as the sole current visual input to
an auxiliary predictor.
We introduce \method{} and its auxiliary objective, the
Action-Conditioned Geometric Joint-Embedding Predictive Architecture
(\objective{}), which
predicts geometric features at several horizons from the current observation
and intervening actions, using the future slot of a frozen VGGT encoding
of each current and future image pair as the target.
We apply this supervision from the head and wrist cameras to a shared
visual embedding before temporal mixing, and remove the teacher and
auxiliary modules at inference.
On 50 RoboTwin~2.0 tasks, \method{} achieves 93.46\% success in clean
scenes, with the best randomized success (92.68\%) and mean across both settings
(93.07\%) among the compared methods; across three tasks on a real robot,
it achieves 85.00\% success and 91.67\% partial completion score, exceeding
Motus by 10.00 and 9.17 percentage points, respectively.
Code: \url{https://github.com/RoboOpus/ACG-WAM}.
Website: \url{https://RoboOpus.github.io/ACG-WAM}.
\end{abstract}

\section{Introduction}
Robot manipulation requires actions that change the scene in a useful way:
a grasp must secure an object, and a placement must leave it in the intended
position. Generative policies learn action sequences from images and language
instructions~\cite{chi2023diffusionpolicy,zitkovich2023rt2,kim2025openvla,black2026pi0visionlanguageactionflowmodel}.
World action models (WAMs) also predict future observations, coupling visual
forecasting with action generation~\cite{bi2026motus,li2026causal,zhu2025uwm}.
This joint formulation gives policy training access to how objects and
robot hands move between observations, in addition to the demonstrated
control commands. For manipulation, the relevant change is often a spatial
relation: a hand closes around an object, a container receives it, or one
object becomes supported by another. Spatial representations have proved
useful for such tasks~\cite{ze2024dp3,goyal2023rvt,ke2025diffuseractor}.
In bimanual manipulation, these relations also connect the two arms:
moving an object with one arm changes the conditions for the other's next action.
The question is how to make geometric information a useful training signal
for a WAM while retaining its existing observation and action interface.

Two challenges arise when adding this supervision. First, video and action
losses do not explicitly require a representation of the geometric change
associated with an action interval. From the same observation, a robot can
approach, lift, or move an object; the starting image alone does not specify
which outcome a predictor should learn. An additional prediction target
must therefore describe a future scene in the context of both its starting
observation and the actions that connect them. Even over the same duration,
different action sequences can lead to different stages of an interaction.
The action interval identifies the demonstrated movement, while its duration
specifies how far ahead the predictor should look. Second, the location of the
supervised policy feature matters. During joint video and action training,
attention mixes tokens from current and future frames. A token assigned to
the current frame can consequently contain future visual information after
this mixing. Using that token as the visual input to an auxiliary predictor
would let it use part of the observation it is supposed to predict. For this supervision to benefit action generation, its gradients must
reach the visual features shared with the policy.

These challenges motivate two complementary choices. For the first,
we predict a geometric representation of a future endpoint conditioned on
the intervening action sequence. A frozen teacher jointly encodes the two
images, allowing the future representation to use the current scene as
context. Sampling different endpoints associates the same current
observation with both short interactions and longer action sequences.
This exposes the visual embedding to changes that develop over different
portions of a manipulation sequence, from the initial approach to later
object transport.
For the second, we supervise the visual embedding before the backbone
mixes frames. The student then receives features computed from the current
image alone, while the teacher can use the future image to form a target.
The head camera describes the shared workspace, including the target region
and the relation between the arms; wrist cameras reveal local configurations
of the grippers and manipulated objects. Together, these views provide
spatial evidence about both the overall arrangement and the immediate interaction.
Supervising the shared embedding with these views connects geometric
prediction to the visual input used by the deployed policy.

We implement these choices as \method{}, a Motus policy trained with the
\objective{} objective. Frozen VGGT~\cite{wang2025vggt} encodes the current
and future images jointly within each camera. We retain its future slot,
pool the features from each view, and train a predictor using the current
visual features, demonstrated actions, and a temporal horizon. The auxiliary
loss updates the shared visual embedding alongside the base video and
action objectives. The teacher and predictor are discarded at inference,
so the policy uses its original inputs and action generation path.
Figure~\ref{fig:teaser} shows this connection between training and
deployment. We evaluate the resulting policy on 50 RoboTwin~2.0 tasks and
three tasks on a real robot. It reaches 93.46\% clean success and leads
the compared methods in randomized success (92.68\%) and the mean across
both settings (93.07\%), with 85.00\% mean success on the robot.
Ablations on six simulation tasks examine action conditioning,
temporal horizons, and the geometric target.
Together, these evaluations test whether geometric endpoint supervision
benefits task completion under visual variation and execution on physical hardware.

In summary, our main contributions are:
\begin{itemize}
\item Geometric dynamics conditioned on actions. We introduce a joint
embedding objective that predicts geometric targets over several horizons
from the current observation and intervening actions. A frozen teacher
jointly encodes current and future images to provide explicit supervision
for world action modeling.
\item Geometric distillation from multiple views. We connect targets from
the head and wrist cameras to policy features computed from the current
observation before temporal mixing. Auxiliary gradients update the shared
visual embedding, while the teacher and predictor are removed at deployment.
\item Benchmark and real robot evaluation. We evaluate the policy on
50 RoboTwin~2.0 tasks and three manipulation tasks on a real robot.
It achieves the best randomized and mean success among the compared
methods on RoboTwin~2.0 and the highest average SR and PCS on the robot.
Ablations on six simulation tasks support the Joint target design,
action conditioning, and supervision at multiple horizons.
\end{itemize}

\begin{figure*}[t]
\centering
\setlength{\abovecaptionskip}{4pt}
\includegraphics[width=\textwidth,trim=56bp 67bp 56bp 59bp,clip]{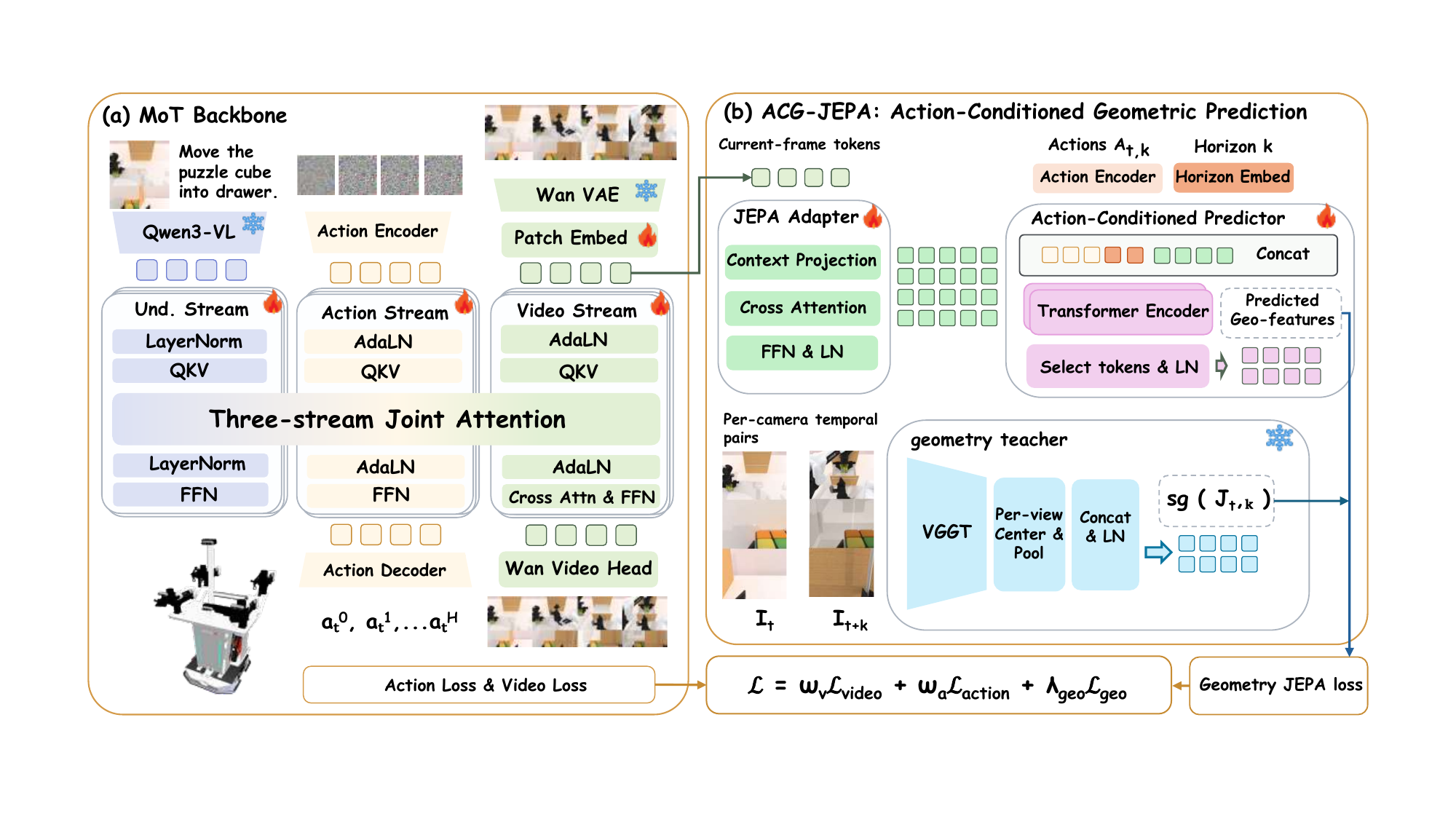}
\caption{Training architecture of ACG-WAM. Motus predicts video and actions.
The auxiliary predictor uses features of the current frame, demonstrated actions,
and a horizon to predict the geometric target. Future images provide teacher
targets without entering the student's visual input. The auxiliary loss updates
the shared patch embedding; the base losses train the backbone.}
\label{fig:architecture}
\end{figure*}

\section{Related Work}
\label{sec:related}
\subsection{Generative Policies and World Action Models}
Robot policies combine pretrained representations of images and language
with action learning. RT-2 and OpenVLA use different approaches to
transfer~\cite{zitkovich2023rt2,kim2025openvla};
$\pi_0$ and $\pi_{0.5}$ focus on continuous control and adaptation~\cite{black2026pi0visionlanguageactionflowmodel,intelligence2025pi05visionlanguageactionmodelopenworld}.
UniPi uses video generation for planning, and GR-1 predicts video and
actions~\cite{du2023unipi,wu2024gr1}.
Unified World Models couple video and action diffusion~\cite{zhu2025uwm}.
Motus combines understanding, generation, and control in a latent WAM,
while LingBot-VA studies causal world modeling for robot
control~\cite{bi2026motus,li2026causal}. We use Motus as the backbone and
add geometric supervision during training.

\subsection{Predictive Representations and Alignment}
Joint embedding methods predict features instead of reconstructing pixels.
I-JEPA and V-JEPA study prediction in learned image and video feature
spaces~\cite{assran2023ijepa,bardes2024vjepa}; V-JEPA~2 extends this
approach to world modeling and planning~\cite{assran2025vjepa2}.
DINO-WM learns dynamics from pretrained visual
features~\cite{zhou2025dinowm}, and JEPA-WAM predicts a joint representation
of current and future observations in a pretrained V-JEPA
space~\cite{lin2026jepawam}.
REPA aligns diffusion representations with features from an external
encoder~\cite{yu2025repa}.
We use joint prediction to supervise geometric features from VGGT.
The predictor connects a demonstrated action interval to its geometric
endpoint, while supervision before temporal attention directly trains
the shared embedding of the current observation. This connects latent
prediction to the visual representation used for action generation.

\subsection{Geometric Representations for Manipulation}
DUSt3R and MASt3R recover spatial structure and correspondences; VGGT
predicts several geometric quantities with a feed-forward
transformer~\cite{wang2024dust3r,leroy2024mast3r,wang2025vggt}.
RVT, 3D Diffusion Policy (DP3), and 3D Diffuser Actor use spatial
representations for control~\cite{goyal2023rvt,ze2024dp3,ke2025diffuseractor}.
WAM4D introduces spatial registers to predict future depth during training,
and MECo-WAM uses a frozen VGGT teacher, action-aware relational
distillation, and restricted expert visibility~\cite{li2026wam4d,zhang2026mecowam}.
Both remove their auxiliary geometry components at inference.
We formulate supervision as geometric endpoint prediction from
the current observation, an intervening action interval, and its horizon.
Targets jointly encode current and future images within each camera,
and supervision at multiple horizons trains the shared visual embedding
used by the deployed policy before temporal attention.

\section{Method}
\label{sec:method}
\subsection{Overview}
\method{} augments Motus with an auxiliary task that predicts geometric
features from the current observation and a sequence of actions.
At time $t$, the policy receives camera observations
$\{I_t^v\}_{v\in\mathcal V}$, robot state $\mathbf s_t$, and a language
instruction $\ell$, and generates a chunk of $H$ actions.
The Motus backbone~\cite{bi2026motus} uses Mixture-of-Transformers (MoT)
attention to couple visual understanding, latent video
generation~\cite{peebles2023dit}, and action generation.
Video and action flow matching~\cite{lipman2023flowmatching} define the
base loss:
\begin{equation}
\mathcal L_{\mathrm{base}} =
w_v\mathcal L_{\mathrm{video}}+w_a\mathcal L_{\mathrm{action}}.
\label{eq:base}
\end{equation}

For the auxiliary task, let $J_{t,k}$ denote a frozen geometric
representation computed jointly from the current and future observations
across the cameras. Let $Z_t$ denote the policy feature computed from the
current observation. The predictor learns
\begin{equation}
(Z_t,\mathbf A_{t,k},k)\longmapsto J_{t,k},
\label{eq:task}
\end{equation}
Here $t$ and $k$ index the downsampled video sequence. Let $r$ be the
number of sampled actions per video transition and $\mathbf a_t^{(i)}$
the $i$th action in the chunk starting at $I_t$, indexed from zero.
Then $\mathbf A_{t,k}=(\mathbf a_t^{(0)},\ldots,\mathbf a_t^{(kr-1)})$
contains exactly $kr$ actions aligned with $I_{t+k}$.
With raw-frame start $f_t$ and action sampling stride $s$, action $i$
is sampled at $f_t+(i+1)s$, and the target image at $f_t+krs$.
This alignment supplies the predictor with the action prefix for each
selected geometric endpoint.
Sec.~\ref{sec:prediction} defines the prediction objective, and
Sec.~\ref{sec:distillation} describes the teacher targets and student features.

\subsection{Geometric Prediction Conditioned on Actions}
\label{sec:prediction}
We form the target by feeding the current image and a future image to a
frozen geometric teacher in a single forward pass. Its attention allows
the representation of the future image to incorporate information from
the current scene. We retain the features at the future slot, giving the
representation $J_{t,k}$ defined in Sec.~\ref{sec:distillation}. The target is
\begin{equation}
D_{t,k}=\sg\!\left(J_{t,k}\right),
\label{eq:joint}
\end{equation}
where $\sg$ stops gradients through the teacher.
Within this auxiliary task, future images are used to construct the target;
the student's visual input is computed from the current observation alone.

Selecting the future slot gives the predictor a geometric endpoint
represented in the context of the initial scene. Joint teacher attention
allows features at that endpoint to incorporate both observations, so
the supervision connects the starting configuration to the demonstrated
outcome in a shared latent representation.

The predictor also receives the actions connecting the observations.
From the same scene, the robot may approach, lift, or move an object,
depending on the action sequence. We combine an adapted visual feature $A_\eta(Z_t)$, an action
embedding $E_a(\mathbf A_{t,k})$, and a horizon embedding $e_k$:
\begin{equation}
Q_{t,k}=g_\theta\!\left(
A_\eta(Z_t),E_a(\mathbf A_{t,k}),e_k
\right).
\label{eq:predictor}
\end{equation}
The action encoder flattens the sequence in time-major order, adaptively
average-pools the resulting vector to a fixed width, and applies two linear
layers with a GELU activation and layer normalization to obtain one token.
The fixed temporal layout gives each pooling bin a consistent position
within the action sequence. A transformer processes this action token,
a learned horizon token, and queries derived from the current visual
features, combining the action interval with the observed scene. Its normalized query outputs form
$Q_{t,k}$. The original policy action branch continues to learn from the
language instruction and the base loss.

During training, one horizon is sampled uniformly for each batch and
shared by its examples. Each future endpoint is paired with the
demonstrated action prefix spanning the same interval. The horizon embedding distinguishes
short interactions from the consequences of longer action sequences. All
horizons use the same predictor, target dimensions, and loss.

For $Q=Q_{t,k}$ and $D=D_{t,k}\in\mathbb R^{M\times d}$, let
$\bar Q$ and $\bar D$ be their means over the $M$ target slots. We use
\begin{align}
\mathcal L_{\mathrm{geo}}={}&
\alpha[1-\cos_\epsilon(\bar Q,\bar D)]
+\beta\frac{\|Q-D\|_F^2}{Md}\nonumber\\
&+\frac{\gamma}{M}\sum_{j=1}^{M}
[1-\cos_\epsilon(Q_j,D_j)],
\label{eq:geo}
\end{align}
where $\cos_\epsilon$ is stabilized cosine similarity, and $\alpha$,
$\beta$, and $\gamma$ weight the three terms.

The loss supervises the representation at two levels: pooled cosine
similarity aligns the overall feature direction, while squared error
and slotwise cosine similarity match individual feature slots. Squared
error penalizes coordinate differences after normalization, and slotwise
cosine similarity aligns the direction of each predicted slot with its
teacher target. Teacher normalization uses fixed scale and bias; the
predictor's output normalization learns these parameters per channel.

\begin{table*}[t]
\centering
\caption{Success rates (\%) on RoboTwin~2.0. Bold marks the best result in each
column of (a) and each method pair in (b). $\Delta$ is in percentage points.}
\label{tab:main}
\normalsize
\setlength{\tabcolsep}{3pt}
\renewcommand{\arraystretch}{1.05}
\begin{minipage}[t]{0.38\textwidth}
\centering
(a) Overall performance on 50 tasks\par\smallskip
\begin{tabular*}{\linewidth}{@{\extracolsep{\fill}}lccc@{}}
\toprule
\multirow{2}{*}{Method} & \multicolumn{3}{c}{RoboTwin 2.0}\\
\cmidrule(lr){2-4}
& Clean & Random & Mean\\
\midrule
GO-1~\cite{bu2025agibot} & 37.80 & 36.24 & 37.02\\
$\pi_{0.5}$~\cite{intelligence2025pi05visionlanguageactionmodelopenworld} & 42.98 & 43.84 & 43.41\\
X-VLA~\cite{zheng2026xvla} & 72.80 & 72.84 & 72.82\\
Motus~\cite{bi2026motus} & 88.66 & 87.02 & 87.84\\
LingBot-VA~\cite{li2026causal} & 91.99 & 91.11 & 91.55\\
MECo-WAM~\cite{zhang2026mecowam} & 93.26 & 91.98 & 92.62\\
WAM4D~\cite{li2026wam4d} & \textbf{93.82} & 89.86 & 91.84\\
\textbf{\method{} (ours)} & 93.46 & \textbf{92.68} & \textbf{93.07}\\
\bottomrule
\end{tabular*}
\end{minipage}\hfill
\begin{minipage}[t]{0.60\textwidth}
\centering
(b) Selected task improvements over Motus\par\smallskip
\setlength{\tabcolsep}{1.5pt}
\renewcommand{\arraystretch}{1.31}
\begin{tabular*}{\linewidth}{@{\extracolsep{\fill}}lcccccc@{}}
\toprule
\multirow{2}{*}{Task} & \multicolumn{3}{c}{Clean} & \multicolumn{3}{c}{Random}\\
\cmidrule(lr){2-4}\cmidrule(lr){5-7}
& Motus~\cite{bi2026motus} & ACG-WAM & $\Delta$
& Motus~\cite{bi2026motus} & ACG-WAM & $\Delta$\\
\midrule
Hanging Mug & 38 & \textbf{70} & +32 & 38 & \textbf{76} & +38\\
Handover Mic & 78 & \textbf{100} & +22 & 63 & \textbf{98} & +35\\
Move Can Pot & 34 & \textbf{98} & +64 & 74 & \textbf{90} & +16\\
Place Mouse Pad & 66 & \textbf{94} & +28 & 68 & \textbf{81} & +13\\
Scan Object & 67 & \textbf{92} & +25 & 66 & \textbf{87} & +21\\
Open Microwave & 95 & \textbf{98} & +3 & 91 & \textbf{99} & +8\\
\bottomrule
\end{tabular*}
\end{minipage}
\end{table*}

\subsection{Geometric Supervision of the Current Observation}
\label{sec:distillation}
The teacher provides targets from three cameras, and the student predictor
reads the shared visual embedding before MoT attention. This placement
lets the geometric loss update a feature extractor used by the deployed policy.

\paragraph{Targets from multiple cameras}
The head camera captures scene layout and the relation between the arms;
the wrist cameras capture local interactions with objects. We process the
head, left wrist, and right wrist views separately,
$\mathcal V=\{h,l,r\}$. For each camera, frozen
VGGT~\cite{wang2025vggt} receives the current image followed by the future image:
\begin{equation}
(X_{t\mid t,k}^{v},X_{t+k\mid t,k}^{v})
=\Psi([I_t^v,I_{t+k}^v]).
\label{eq:pair}
\end{equation}
We use the deepest available VGGT aggregator output and select its
future slot $X_{t+k\mid t,k}^{v}$, which can attend to both images.
We abbreviate it as $X_{t,k}^{v}$. Each channel is centered across
the view's $N$ tokens:
\begin{equation}
\widetilde X_{t,k,n}^v =
X_{t,k,n}^v-\frac{1}{N}\sum_{j=1}^{N}X_{t,k,j}^v.
\label{eq:center}
\end{equation}
Adaptive average pooling $P$ first reduces the channel axis to $d=768$
and then the token axis to 32 slots per camera, without learned parameters.
We concatenate the views in head, left wrist, and right wrist order and
apply layer normalization independently over each slot's feature axis,
without learned scale or bias:
\begin{equation}
J_{t,k}=\LN\!\left(
[P(\widetilde X_{t,k}^h);P(\widetilde X_{t,k}^l);P(\widetilde X_{t,k}^r)]
\right)\in\mathbb R^{M\times d}.
\label{eq:teacher}
\end{equation}
Here $M$ is the total number of pooled slots and $d$ is their feature width.
The teacher mixes temporal information within each camera pair. Separate
slots retain the identity of each view after concatenation. No subtraction
of independently encoded endpoints is used in this target.

\begin{figure}[!t]
\centering
\includegraphics[width=\columnwidth,trim=79bp 182bp 244bp 142bp,clip]{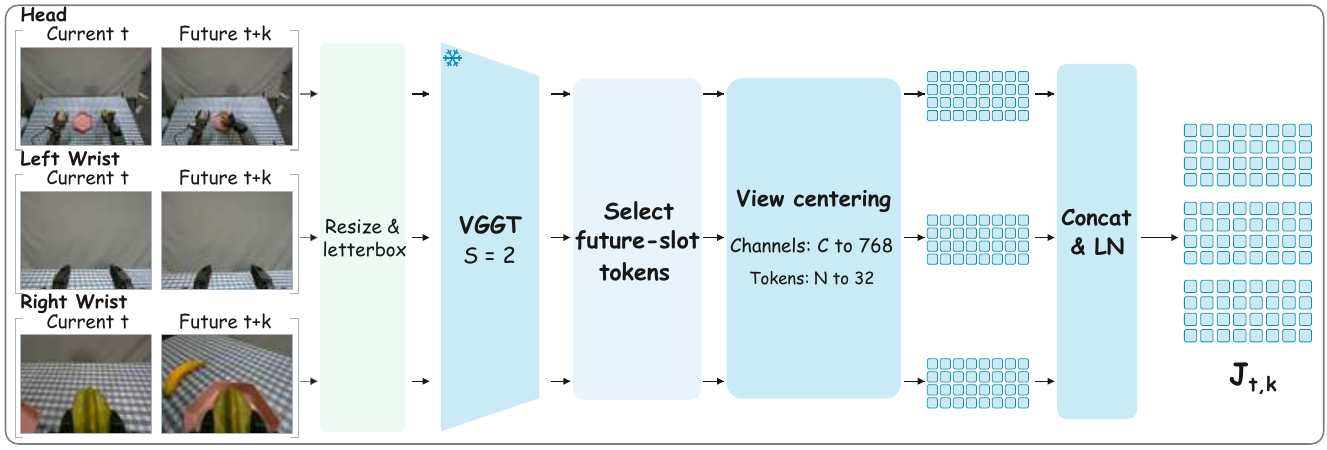}
\caption{Construction of the geometric target with frozen VGGT.
Each camera contributes an ordered pair of current and future images.
The future slot is centered and pooled within each view before the three
views are concatenated and normalized. The target branch is used only
during training.}
\label{fig:teacher}
\end{figure}

Figure~\ref{fig:teacher} details this construction. In our implementation,
pooling produces 32 slots of width 768 for each camera, giving
$M=96$ after concatenation. Centering removes the mean across tokens
within each view, and pooling produces the fixed number of target slots.

\paragraph{Features of the current observation}
Future video latents participate in the backbone's denoising loss.
After attention mixes frames and streams, even tokens assigned to the
current frame can contain future visual information. We therefore read
the student feature before the first MoT layer.

The frozen video VAE, $\mathcal E$, encodes the current image separately.
Its latent serves as the clean conditioning frame. The shared patch
embedding $F_\phi$, applied to each frame, produces
\begin{equation}
Z_t=F_\phi\!\left(\mathcal E(I_t)\right).
\label{eq:student}
\end{equation}
In the joint computation, $Z_t$ is the slice corresponding to the current
frame. Separate encoding and a patch embedding with no temporal mixing
ensure that it depends on $I_t$ alone. The auxiliary predictor receives
this feature, the demonstrated actions, and the horizon. Future video
latents remain inputs to the base training objective.

The adapter $A_\eta$ pools and projects the input channels, adds learned
spatial embeddings, and resamples the patches using cross-attention from
learned queries. A residual feed-forward block and normalization complete
the adapter. Its fixed set of output queries predicts the teacher slots
while attending to spatially indexed features of the current frame.

\paragraph{Training and inference}
The total loss is
\begin{equation}
\mathcal L=\mathcal L_{\mathrm{base}}+
\lambda_{\mathrm{geo}}\mathcal L_{\mathrm{geo}}.
\label{eq:total}
\end{equation}
The geometric loss updates the adapter, action encoder, predictor, and
shared patch embedding $F_\phi$. The embedding also supplies the video
stream, so it receives gradients from both $\mathcal L_{\mathrm{base}}$
and $\mathcal L_{\mathrm{geo}}$. The teacher and VAE remain frozen.
Downstream MoT layers and the action decoder are optimized by the base
loss using the updated embedding. This shared parameter path carries
the auxiliary supervision into the policy.

At inference, the policy uses the trained Motus backbone. The teacher,
target cache, adapter, and predictor are removed; the shared visual
embedding remains. Geometric supervision therefore adds no module to
the deployed computation graph.

\section{Experiments}
\label{sec:protocol}
We evaluate \method{} on RoboTwin~2.0 and on three tasks with a real
robot. The main comparisons measure task success, and the ablations
compare prediction targets, action conditioning, and temporal horizons.

\subsection{Implementation}
\paragraph{Model implementation}
The backbone uses Wan2.2-TI2V-5B and Qwen3-VL-2B-Instruct
~\cite{wanvideo2025wan22,bai2025qwen3vl}. Frozen VGGT targets are cached
offline following Sec.~\ref{sec:distillation}, resizing wrist crops to the
head-view size before preprocessing. The 96-query adapter and predictor
have one and two blocks, respectively, with width 768, eight heads,
feed-forward width 3072, and zero dropout. The action encoder has two
$768\times768$ linear layers. Geometric losses use float32.

\paragraph{Training}
We initialize \method{} from the pretrained base model and train on
RoboTwin for 40k updates with \objective{} enabled throughout.
This matches the budget reported in Motus~\cite{bi2026motus}; LingBot-VA
uses 50k updates~\cite{li2026causal}. We use joint future-slot targets,
$k\in\{1,2,4,8\}$, $r=2$, and $s=3$: the eight-transition window
contains $H=16$ actions, with six raw frames per video transition.
AdamW optimizes Eq.~\eqref{eq:total} with base learning rate
$5\times10^{-5}$, weight decay $0.01$, and moments $(0.9,0.95)$.
After 200 warmup updates, the learning-rate multiplier decays linearly
from $0.99$ to $0.4$. Four GPUs with microbatch four and 16-step gradient
accumulation give a global batch of 256. We use gradient clipping $0.5$
and bfloat16 mixed precision. Loss weights are $\lambda_{\mathrm{geo}}=0.01$
and $(\alpha,\beta,\gamma)=(1,0.05,1)$, with unit weights for the base losses.

\paragraph{RoboTwin inference}
We use ten UniPC steps with shift $1.0$ for video and action generation,
executing all 16 joint-position actions before the next policy observation.

\subsection{Simulation Experiments}
\paragraph{Setting}
We evaluate on RoboTwin~2.0~\cite{chen2025robotwin2}, which comprises
50 bimanual manipulation tasks. The \emph{demo-clean} and
\emph{demo-randomized} settings test nominal scenes and visually randomized
environments, respectively. The task suite spans object relocation,
stacking, tool interaction, and coordinated bimanual manipulation.

\begin{figure}[!ht]
\centering
\includegraphics[width=\columnwidth]{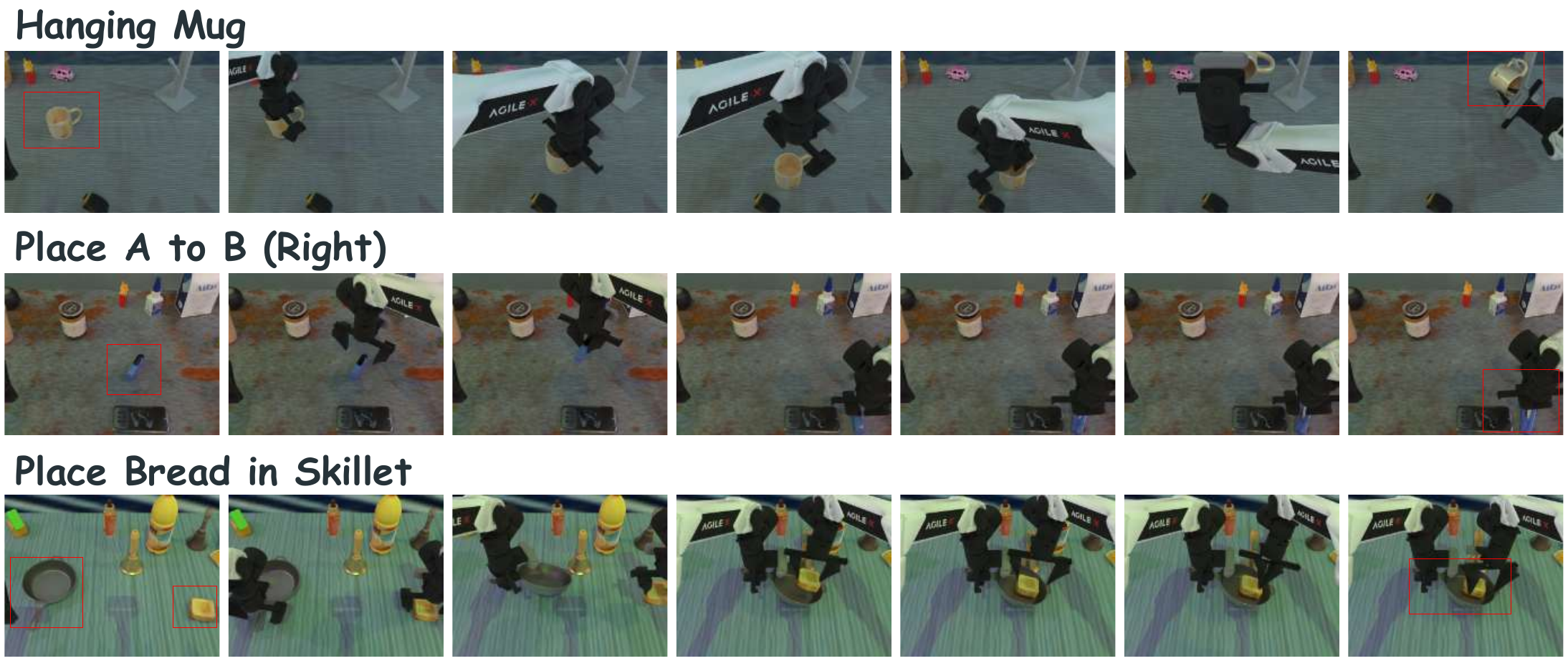}
\caption{Successful ACG-WAM trajectories for Hanging Mug (top),
Place A to B (Right) (middle), and Place Bread in Skillet (bottom)
in randomized RoboTwin~2.0 scenes. Red boxes mark the manipulated objects
in the first frame of each sequence.}
\label{fig:simulation}
\end{figure}

\begin{figure*}[!t]
\centering
\includegraphics[width=\textwidth,trim=7bp 124bp 35bp 5bp,clip]{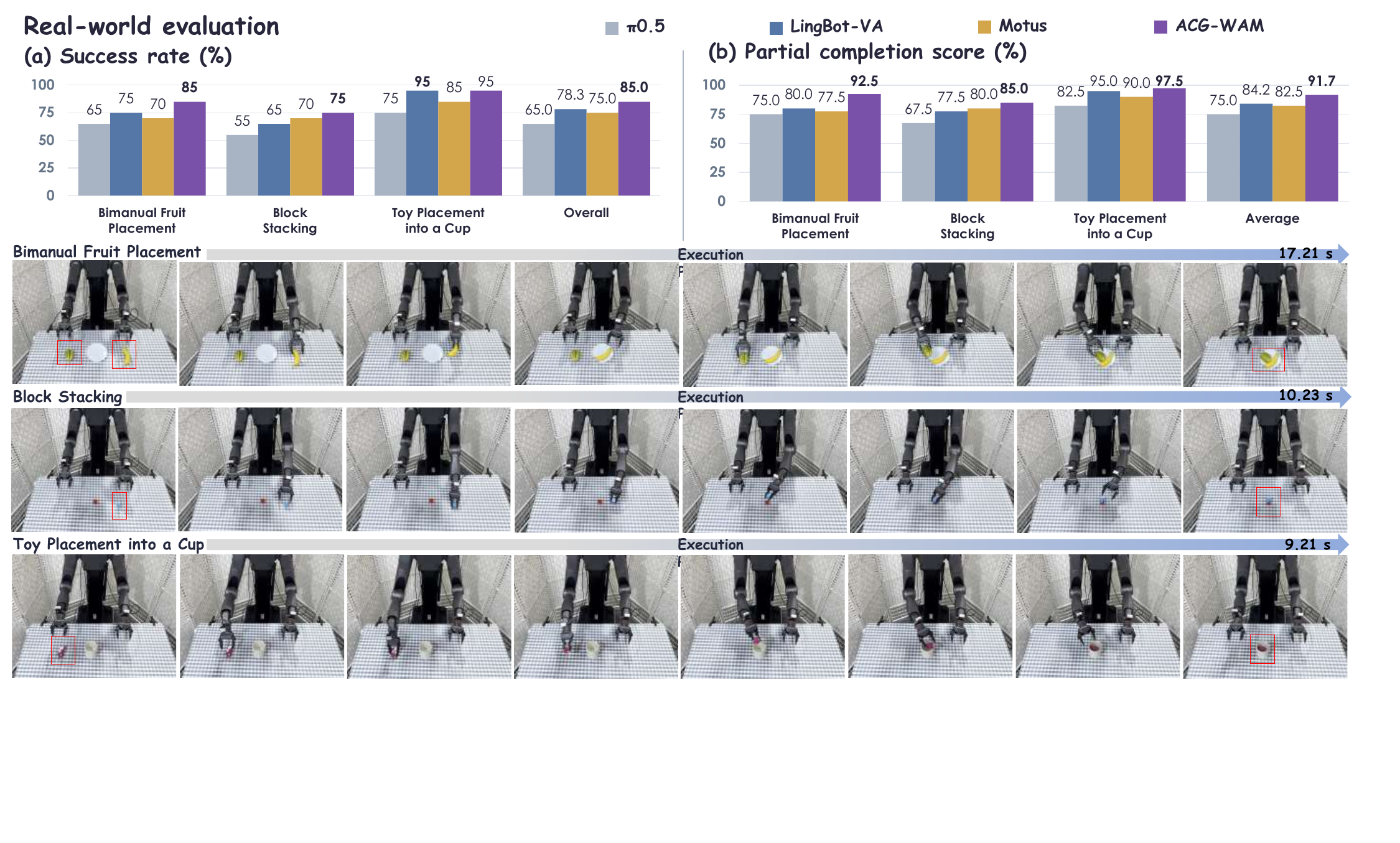}
\caption{Evaluation on a real robot across three manipulation tasks.
The upper panels show success rate and partial completion score.
Below, eight keyframes per task show execution from left to right;
arrows indicate elapsed time.}
\label{fig:realworld}
\label{fig:platform}
\end{figure*}

\paragraph{Baselines}
Table~\ref{tab:main}(a) compares \method{} with GO-1,
$\pi_{0.5}$, X-VLA, Motus, LingBot-VA, MECo-WAM, and WAM4D.
Results for the first four methods are taken from Motus~\cite{bi2026motus};
the remaining baseline results are from their respective
publications~\cite{li2026causal,zhang2026mecowam,li2026wam4d}.

\paragraph{Evaluation}
For \method{}, each task is evaluated in each setting over 100 episodes with unseen
instruction templates and evaluation seed 42. Success follows the
benchmark's completion criterion for each task. Table~\ref{tab:main}
reports macro averages over all 50 tasks for each setting and six task
comparisons with Motus. Mean gives equal weight to Clean and Random.

\paragraph{Results}
\method{} achieves 93.46\% clean success and the highest randomized
success (92.68\%) and mean across both settings (93.07\%) among the compared
methods in Table~\ref{tab:main}(a). It exceeds MECo-WAM in both settings
and achieves clean performance close to WAM4D, with a 2.82-point
advantage under randomization. Relative to Motus, geometric supervision
improves clean and randomized success by 4.80 and 5.66 percentage points.

Across all 50 tasks, the clean and randomized win/tie/loss counts
relative to Motus are 31/7/12 and 37/5/8, respectively.
Table~\ref{tab:main}(b) illustrates these gains across distinct interactions,
including hanging, handover, relocation, scanning, and opening an
articulated object. The randomized gains reach 38 points on
\emph{Hanging Mug} and 35 points on \emph{Handover Mic}.
The ablations in Sec.~\ref{sec:ablations} compare geometric targets,
action conditioning, and temporal horizons.

\paragraph{Qualitative examples}
Figure~\ref{fig:simulation} shows successful randomized episodes of
\emph{Hanging Mug}, \emph{Place A to B (Right)}, and
\emph{Place Bread in Skillet}. The sequences illustrate completion of
three spatial goals: supporting a mug on a rack, positioning one object
relative to another, and placing bread within a receptacle.

\begin{table*}[t]
\centering
\caption{Success rate (SR) and partial completion score (PCS) on the real robot (\%).
Best results in each column are bold.}
\label{tab:robot}
\small
\setlength{\tabcolsep}{5pt}
\renewcommand{\arraystretch}{1.15}
\begin{tabular*}{\textwidth}{@{\extracolsep{\fill}}l*{8}{c}@{}}
\toprule
\multirow{2}{*}{Model}
& \multicolumn{2}{c}{\shortstack{Bimanual Fruit\\Placement}}
& \multicolumn{2}{c}{\shortstack{Block\\Stacking}}
& \multicolumn{2}{c}{\shortstack{Toy Placement\\into a Cup}}
& \multicolumn{2}{c}{\shortstack{Task\\Average}}\\
\cmidrule(lr){2-3}\cmidrule(lr){4-5}\cmidrule(lr){6-7}\cmidrule(lr){8-9}
& SR & PCS & SR & PCS & SR & PCS & SR & PCS\\
\midrule
$\pi_{0.5}$~\cite{intelligence2025pi05visionlanguageactionmodelopenworld} & 65 & 75.0 & 55 & 67.5 & 75 & 82.5 & 65.00 & 75.00\\
LingBot-VA~\cite{li2026causal} & 75 & 80.0 & 65 & 77.5 & \textbf{95} & 95.0 & 78.33 & 84.17\\
Motus~\cite{bi2026motus} & 70 & 77.5 & 70 & 80.0 & 85 & 90.0 & 75.00 & 82.50\\
\method{} (ours) & \textbf{85} & \textbf{92.5} & \textbf{75} & \textbf{85.0} & \textbf{95} & \textbf{97.5} & \textbf{85.00} & \textbf{91.67}\\
\bottomrule
\end{tabular*}
\end{table*}

\subsection{Experiments on a Real Robot}
\label{sec:realworld}
\paragraph{Platform and tasks}
We evaluate bimanual manipulation on the TRON2 platform equipped with
WUJI hands. A head camera observes the workspace,
and two wrist cameras capture local interactions with objects.
The task suite consists of \emph{Bimanual Fruit Placement},
\emph{Block Stacking}, and \emph{Toy Placement into a Cup}.
Together, these tasks cover coordinated object placement, stable stacking,
and placement within a confined receptacle. Figure~\ref{fig:realworld}
shows the evaluation overview and example task executions. We compare \method{} with
$\pi_{0.5}$~\cite{intelligence2025pi05visionlanguageactionmodelopenworld},
LingBot-VA~\cite{li2026causal}, and Motus~\cite{bi2026motus}
under a common protocol for demonstrations on this platform,
observation preprocessing, and action representation.
We collect 100 demonstration episodes per task and fine-tune the policies
for 15k updates on the corresponding task data.

\paragraph{Evaluation}
Each policy is evaluated over 20 trials per task, giving 60 trials per
policy. We report task success rate (SR) and a normalized
partial completion score (PCS) to distinguish complete execution from
partial progress. Each trial receives a score $s_{j,i}\in\{0,1,2\}$:
zero for no completion, one for partial completion, and two for full
completion. For $N_j$ trials on task $j$,
\begin{equation}
\mathrm{SR}_j=\frac{100}{N_j}\sum_i\mathbf{1}[s_{j,i}=2],
\qquad
\mathrm{PCS}_j=\frac{100}{2N_j}\sum_i s_{j,i}.
\label{eq:progress}
\end{equation}
For \emph{Bimanual Fruit Placement}, the scores count zero, one, or two
fruits stably placed in the target area.
For \emph{Block Stacking}, partial credit denotes placing the upper
block before it falls; full credit requires the released stack to remain
stable for at least three seconds.
For \emph{Toy Placement into a Cup}, partial credit denotes lifting the
toy, while full credit requires releasing it into the cup and retaining
it there for at least three seconds.
The reported average gives equal weight to the three tasks.
Table~\ref{tab:robot} pairs SR and PCS for the compared policies.

\paragraph{Results}
\method{} achieves the highest average SR and PCS among the evaluated
policies, reaching 85.00\% and 91.67\%, respectively
(Table~\ref{tab:robot}). These exceed Motus by 10.00 and 9.17 percentage
points, extending the simulation gains to physical bimanual manipulation.
It leads both metrics on \emph{Bimanual Fruit Placement} and
\emph{Block Stacking}. On \emph{Toy Placement into a Cup}, it ties
LingBot-VA at 95\% SR and achieves the highest PCS of 97.5\%.
Together, the SR and PCS gains show improved complete execution and
progress through the prescribed manipulation stages.

\subsection{Ablation Studies}
\label{sec:ablations}
We examine three design choices in \objective{}: the geometric prediction
target, action conditioning, and the temporal horizons used for supervision.
All variants are trained for 8k updates on the same six tasks:
\emph{Handover Mic}, \emph{Hanging Mug}, \emph{Move Can Pot},
\emph{Place Mouse Pad}, \emph{Put Object Cabinet}, and \emph{Scan Object}.
Each task is evaluated over 100 episodes per setting with seed 42.
Tables~\ref{tab:horizons} and~\ref{tab:targets} report average success
rates. The Mean column reports the average of the clean and randomized
success rates. All comparisons share the same full Joint checkpoint.

\paragraph{Effect of prediction horizons}
We compare sampling from $k\in\{1,2,4,8\}$ with a fixed short
horizon ($k=1$) and a fixed long horizon ($k=8$). All three variants
retain the joint target and action conditioning.
Here $k$ counts downsampled video transitions within the training window;
the action prefix is aligned with the selected target endpoint.

\begin{table}[!htb]
\centering
\caption{Prediction horizon ablation on RoboTwin~2.0. Values are success rates (\%).}
\label{tab:horizons}
\small
\setlength{\tabcolsep}{3pt}
\renewcommand{\arraystretch}{1.12}
\begin{tabular*}{\columnwidth}{@{\extracolsep{\fill}}lrrr@{}}
\toprule
Horizon set & Clean & Random & Mean\\
\midrule
Short $\{1\}$ & 79.33 & 79.17 & 79.25\\
Long $\{8\}$ & 82.17 & 80.67 & 81.42\\
Multiple $\{1,2,4,8\}$ & \textbf{84.67} & \textbf{82.67} & \textbf{83.67}\\
\bottomrule
\end{tabular*}
\end{table}

Sampling multiple horizons gives the highest mean success, 83.67\%,
in Table~\ref{tab:horizons}. It exceeds the fixed long and short
horizons by 2.25 and 4.42 percentage points, respectively. The difference is particularly clear
on \emph{Hanging Mug}: sampling multiple horizons reaches 46\% clean and
38\% randomized success, compared with 39\%/28\% for the long horizon and
22\%/20\% for the short horizon.
Supervision at multiple horizons links each current observation to action
intervals of different lengths, covering both short interactions and
longer manipulation sequences.

\paragraph{Effect of the prediction target}
We compare the final Joint target with subtraction of endpoint features.
Both variants use action conditioning, sample the same horizons, and read
student features before temporal mixing. Joint uses the teacher's future
slot in Eq.~\eqref{eq:joint}. Residual encodes each endpoint independently
with the same frozen VGGT and the same pooling for each view, then forms
\begin{equation}
D^{\mathrm{res}}_{t,k}=\sg(U_{t+k}-U_t),
\label{eq:residual}
\end{equation}
where $U_s$ pools and normalizes the features from all camera views
at time $s$. The predictor output shape and loss family
are unchanged.

\begin{table}[!htb]
\centering
\caption{Target and action conditioning ablations on RoboTwin~2.0. Values are success rates (\%).}
\label{tab:targets}
\small
\setlength{\tabcolsep}{3pt}
\renewcommand{\arraystretch}{1.12}
\begin{tabular*}{\columnwidth}{@{\extracolsep{\fill}}lrrr@{}}
\toprule
Variant & Clean & Random & Mean\\
\midrule
Endpoint residual & 79.83 & 81.00 & 80.42\\
Joint, no actions & 83.00 & 80.83 & 81.92\\
Joint (full) & \textbf{84.67} & \textbf{82.67} & \textbf{83.67}\\
\bottomrule
\end{tabular*}
\end{table}

Using the jointly encoded future target increases mean success from 80.42\% to
83.67\%, a gain of 3.25 percentage points.
The clean and randomized gains are 4.83 and 1.67 points, respectively.
Joint improves four tasks in each setting; on \emph{Hanging Mug},
success rises from 30\%/32\% to 46\%/38\%, while on
\emph{Place Mouse Pad} it rises from 82\%/80\% to 89\%/84\%.
These results show the advantage of the Joint target design over
endpoint subtraction for geometric supervision. Joint teacher attention
constructs the future representation with access to the initial scene,
providing an endpoint target that carries temporal context.

\paragraph{Effect of action conditioning}
We replace the geometric predictor's action sequence with zeros before
the action encoder, retaining the Joint target, horizon sampling,
model architecture, and policy loss. In Table~\ref{tab:targets}, action
conditioning increases mean success from 81.92\% to 83.67\%, a gain of
1.75 percentage points, or 21 additional successes over 1,200 episodes.
Clean and randomized success improve by 1.67 and 1.83 points, respectively.
The largest gain occurs on \emph{Hanging Mug}, where randomized success
rises from 27\% to 38\%.
These results support conditioning geometric prediction on the intervening
action sequence: actions specify how the current scene evolves, making
geometric supervision more effective for manipulation.

\section{Conclusion}
We presented \method{} to give world action models explicit supervision
for the geometric consequences of robot actions. Its \objective{}
objective predicts jointly encoded geometric endpoints from the current
observation and intervening actions at several horizons. Supervision
from the head and wrist cameras updates the policy's shared visual
embedding before temporal mixing.
This supervision lets the visual embedding learn from the geometry of
completed transitions while remaining computable from the robot's current
observations.

Evaluations on RoboTwin~2.0 and three real robot tasks demonstrate the
value of this supervision for manipulation. \method{} leads the compared
methods in randomized and mean benchmark success and achieves the highest
average SR and PCS on the real robot. The ablations support the Joint
target design, action conditioning, and prediction at several horizons,
showing how these choices contribute to the final policy's success.
Together, these results establish geometric endpoint prediction as an
effective way to connect world modeling with action generation. The
resulting policy benefits from geometric supervision while retaining
the backbone's observation interface and computation at deployment.

\section*{Acknowledgment}
ChatGPT was used to assist with modifications to the experimental code
and with drafting and polishing the Related Work and Method sections.
All generated outputs were reviewed, edited, and fact-checked by the authors.

\bibliographystyle{IEEEtran}
\bibliography{references}
\end{document}